\documentclass[10pt,twocolumn,letterpaper]{article}

\usepackage{cvpr}              % To produce the CAMERA-READY version
\usepackage{graphicx}
\usepackage{amsmath}
\usepackage{amssymb}
\usepackage{booktabs}
\usepackage{paralist}

\usepackage{multirow}
\usepackage{multicol}
\usepackage{tabularx}
\usepackage{makecell}
\usepackage[normalem]{ulem}

\usepackage{xcolor,colortbl}
\definecolor{mygray}{rgb}{.9,.9,.9}
\definecolor{mygray1}{rgb}{.9,.9,.98}

\usepackage[pagebackref,breaklinks,colorlinks]{hyperref}

\usepackage[capitalize]{cleveref}
\crefname{section}{Sec.}{Secs.}
\Crefname{section}{Section}{Sections}
\Crefname{table}{Table}{Tables}
\crefname{table}{Tab.}{Tabs.}

\newcommand{\ig}{\textit{i}.\textit{e}.}
\usepackage[symbol]{footmisc}

\def\cvprPaperID{*****} % *** Enter the CVPR Paper ID here
\def\confName{CVPR}
\def\confYear{2022}

\begin{document}

\title{Neural Architecture Search on Efficient Transformers and Beyond}

\author{Zexiang Liu$^{1}$, Dong Li$^{1}$, Kaiyue Lu$^{1}$, Zhen Qin$^{1}$, Weixuan Sun$^{3,1}$, Jiacheng Xu$^{1}$, Yiran Zhong$^{1,2}$\thanks{Zexiang Liu, Dong Li, and Kaiyue Lu contributed equally to this research. Corresponding author: Yiran Zhong (\textit{zhongyiran@gmail.com})} \\
\small $^1$SenseTime Research \hspace{2em} \small $^2$Shanghai AI Lab \hspace{2em} \small $^3$Australian National University
}

% \author{\small $^1$SenseTime Research \hspace{2em} \small $^2$Shanghai AI Lab \hspace{2em} \small $^3$Australian National University}

% \authornote{Zexiang Liu, Dong Li, and Kaiyue Lu contributed equally to this research. \\
% *Corresponding author: Yiran Zhong (\textit{zhongyiran@gmail.com})

% \and
\maketitle

%%%%%%%%% ABSTRACT
\begin{abstract}
   Recently, numerous efficient Transformers have been proposed to reduce the quadratic computational complexity of standard Transformers caused by the Softmax attention. However, most of them simply swap Softmax with an efficient attention mechanism without considering the customized architectures specially for the efficient attention. In this paper, we argue that the handcrafted vanilla Transformer architectures for Softmax attention may not be suitable for efficient Transformers. To address this issue, we propose a new framework to find optimal architectures for efficient Transformers with the neural architecture search (NAS) technique. The proposed method is validated on popular machine translation and image classification tasks. We observe that the optimal architecture of the efficient Transformer has the reduced computation compared with that of the standard Transformer, but the general accuracy is less comparable. It indicates that the Softmax attention and efficient attention have their own distinctions but neither of them can simultaneously balance the accuracy and efficiency well. This motivates us to mix the two types of attention to reduce the performance imbalance. Besides the search spaces that commonly used in existing NAS Transformer approaches, we propose a new search space that allows the NAS algorithm to automatically search the attention variants along with architectures. Extensive experiments on WMT'14 En-De and CIFAR-10 demonstrate that our searched architecture maintains comparable accuracy to the standard Transformer with notably improved computational efficiency.  
\end{abstract}

\section{Introduction}
Efficient Transformers~\cite{lee2019set,katharopoulos2020transformers,zhen2022cosformer,sun2022vicinity} have achieved remarkable advances in recent years. They reduce the quadratic computational complexity of the standard Transformer \cite{vaswani2017attention} by spasifying or approximating Softmax attention in a more efficient fashion. Currently, the configuration of the efficient network, \eg, the number of heads and the input embedding size, is directly copied from the Transformer, which may not be suitable for efficient Transformers \cite{zhen2022cosformer}. The customized network structure specifically for efficient Transformers has not been well studied. However, the manual design always involves expensive engineering work and could be sub-optimal due to the human bias \cite{guan2021autoattend}. Hence, in this paper, we aim to explore how to automatically find an appropriate architecture for efficient Transformers.

\begin{figure}[tb]
\centering
\includegraphics[width=\linewidth]{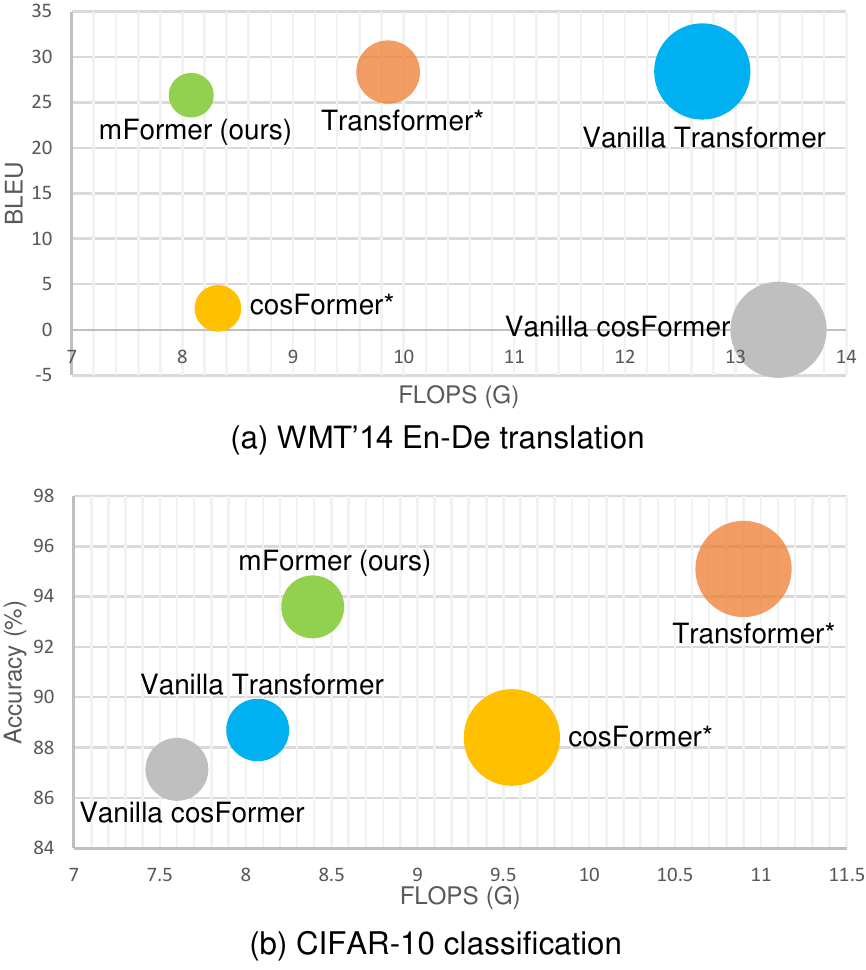}\\
%\end{tabular}
%\vspace{-6pt}
%\vspace{-12pt}
\caption{Performance comparison in accuracy and efficiency. \textnormal{Models marked with ``*'' are the optimal architectures that found by NAS. The size of the circle represents the relative parameter scale on each task. The proposed mixed use of Softmax and efficient attention, \ig, mFormer, is beneficial for balancing accuracy and efficiency. Notice: (a) The cosFormer presents very low BLEU on the translation task due to the inappropriate convergence. (b) The vanilla Transformer and cosFormer in CIFAR-10 have a smaller FLOPS because they only have 6 layers, while the searched optimal models (including ours) have more than 10 layers. With similar FLOPS, ours has a much higher accuracy.}}
\vspace{-15pt}
\label{fig:intro}
\end{figure}

To achieve the goal, we need to (1) specify a suitable efficient Transformer as the target, and (2) find an approach to automating the network design. 
For the first point, we give a brief overview of existing efficient Transformers. 
In terms of how to treat the Softmax attention, they can be roughly classified into pattern-based and kernel-based \cite{zhen2022cosformer}. Pattern-based methods sparsify the attention matrix with predefined or learnable patterns, \eg, chunking input sequences into fixed blocks \cite{qiu2019blockwise}, calculating attention at fixed intervals \cite{child2019generating}, or using axial attention \cite{ho2019axial}. Although the sparsity generated by specific patterns is beneficial for simplifying attention calculation, it still suffers from the quadratic complexity with respect to the input length $N$ \cite{tay2020efficient} and is complicated to implement. 
Differently, kernel-based methods aim at reducing the quadratic complexity into linear (\ig, $O(N)$ in both time and space complexity) \cite{zhen2022cosformer,katharopoulos2020transformers}. They reformulate the self-attention mechanism to avoid explicitly computing the $N\times N$ matrix. Moreover, they are easier to reproduce in practice.
% avoiding the $N\times N$ matrix by reformulating attention, 
% and reduce the quadratic complexity into linear (\ig, $O(N)$ in both time and space complexity) \cite{zhen2022cosformer,katharopoulos2020transformers}. 
Considering the intriguing properties of kernel methods, we choose the cosFormer \cite{zhen2022cosformer}, a new kernel-based model with state-of-the-art performance among efficient Transformers, as our target. We attempt to seek for its customized and optimal architecture.

\begin{figure}[tb]
\centering
\includegraphics[width=\linewidth]{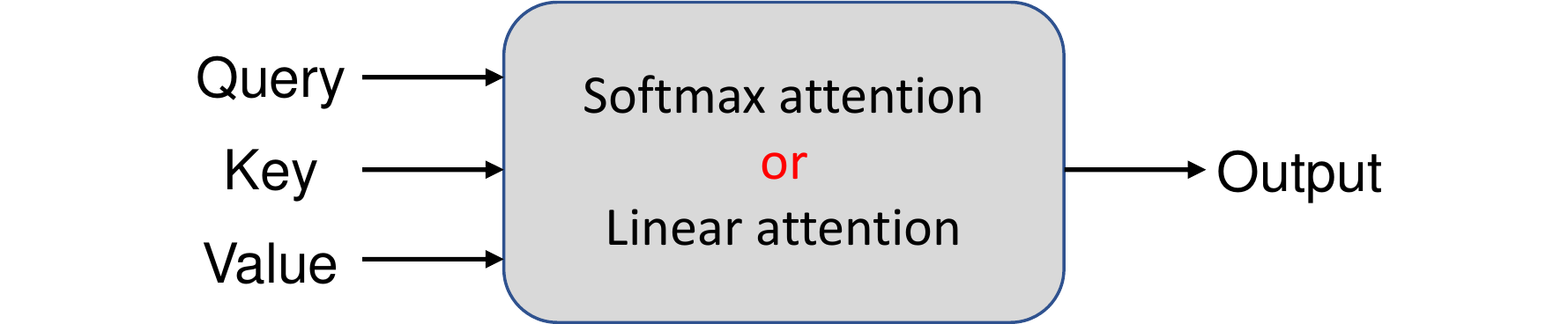}\\
%\end{tabular}
%\vspace{-6pt}
%\vspace{-12pt}
\caption{The proposed search space for attention types. \textnormal{It contains the Softmax attention and linear attention (\ig, cosFormer attention \cite{zhen2022cosformer} in this paper). We let the model automatically determine which attention to use with NAS. }}
\vspace{-7pt}
\label{fig:proposed_attn}
\end{figure}

To automate the network design, we take advantage of neural architecture search (NAS) \cite{elsken2019neural,liu2021survey}. It has been widely employed in searching standard Transformer architectures in Natural Language Processing \cite{so2019evolved,guan2021autoattend,wang2020textnas,ranknas,hat} and Computer Vision \cite{chen2021glit,gong2021nasvit,ding2021hr,chen2021autoformer}. These studies mainly focus on refining search space and/or improving search algorithms. For example, AutoAttend \cite{guan2021autoattend} explores the effecitve connections between the \textit{Query}, \textit{Key}, and \textit{Value} by creating primitive operations to the input data, \eg, $1\times 1$ convolution and $1\times3$ max pooling. The evolved Transformer \cite{so2019evolved} applies tournament selection \cite{real2019regularized} to NAS to generate candidate models more robustly. However, these methods suffer from long training and large search costs because all the candidates need to be optimized, evaluated, and ranked. For the purpose of lowering these costs, we utilize RankNAS \cite{ranknas}, a new efficient NAS framework for searching the standard Transformer \cite{vaswani2017attention}. It can significantly speed up the search procedure through pairwise ranking, search space pruning, and the hardware-aware constraint.

Given both the efficient Transformer (\ig, cosFormer \cite{zhen2022cosformer}) and the search algorithm (\ig, RankNAS \cite{ranknas}), we conduct a preliminary study on pure efficient Transformer-based NAS. Specifically, we replace Softmax with the linear attention introduced in the cosFormer, and then search it with RankNAS. 
To comprehensively investigate the optimal architecture, we perform the search on two representative tasks in NLP and CV fields, \ig, machine translation (WMT'14 En-De \cite{wmt}) and image classification (CIFAR-10 \cite{cifar}). Generally, we observe that the optimal structure of the cosFormer has fewer computational costs, \eg, smaller FLOPS as illustrated in Fig.~\ref{fig:intro}. However, the general accuracy is less comparable to that of the standard Transformer (see the vertical axis in Fig.~\ref{fig:intro}). This performance imbalance between accuracy and efficiency has also been revealed in other efficient Transformers \cite{tay2020sparse,katharopoulos2020transformers,Reformer,wang2020linformer,2020masked}.

Considering that the vanilla Softmax attention and linear attention have their own distinctions regarding the performance, we propose to use Softmax and linear attention in a mixed way for the benefit of a better balance between accuracy and efficiency (named as ``mFormer''). 
Moreover, we expect the model to determine which type of attention to use automatically. To this end, we introduce a new search space specially for attention and incorporate it into the NAS framework. After re-searching the optimal architecture, we find that the combination of the two attention types achieves comparable performance to the Transformer with notably improved efficiency on both tasks (see Fig.~\ref{fig:intro}).

In summary, we make three primary contributions:
\vspace{-3pt}
\begin{compactitem}

    \item To the best of our knowledge, it is the first work to search the optimal architecture of efficient Transformers. We employ NAS for searching the cosFormer, a representative efficient model. The searched results give new insights to the community, \ig, how to design customized, effective, and efficient network structures for efficient Transformers. 
    \item We propose a novel usage of attention, \ig, mixing Softmax attention and linear attention in the Transformer, and define a new search space for attention search in the NAS framework. This pushes existing Transformer search further by enabling automatic selection of appropriate attention types.
    \item The proposed mixed attention achieves better balance between accuracy and efficiency, \ig, having comparable performance to the standard Transformer while maintaining good efficiency. This is validated on both machine translation and image classification tasks.
\end{compactitem}

\begin{figure*}[tb]
\centering
\includegraphics[width=0.9\linewidth]{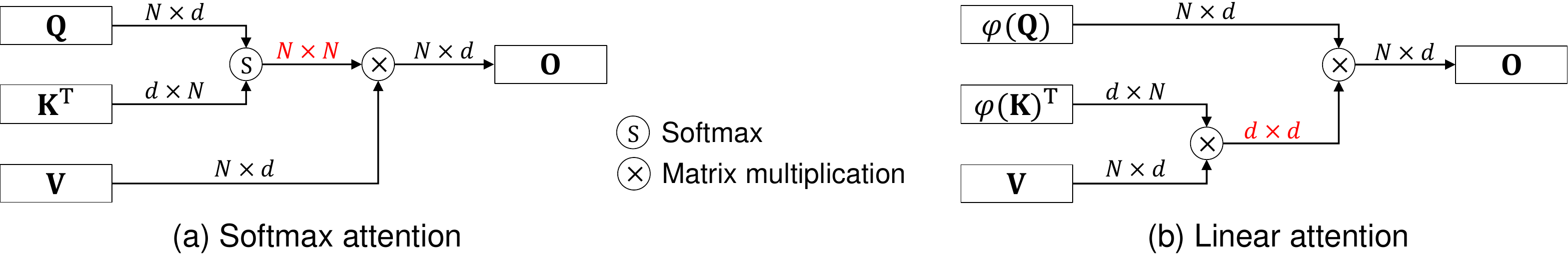}\\
%\end{tabular}
\vspace{-6pt}
%\vspace{-12pt}
\caption{Illustration of attention. \textnormal{(a) Softmax calculates $\mathbf{Q}\mathbf{K}^{\mathrm{T} }$ first, leading to the quadratic complexity with respect to the sequence length $N$. (b) Linear attention decomposes the similarity function with a kernel function, and multiplies $\mathbf{K}$ and $\mathbf{V}$ first. This reduces the quadratic complexity to linear.}}
\vspace{-2pt}
\label{fig:attn_illus}
\end{figure*}

\section{Related Work}
\subsection{Efficient transformers}
The standard Transformer \cite{vaswani2017attention} suffers from the quadratic time and space complexity caused by the Softmax attention. To reduce computational costs, 
% \weixuan{efficient transformers can be generally categorized into pattern based methods and kernel based methods.}
efficient Transformers either sparsify or approximate Softmax in a more efficient fashion. 
Sparsifying the attention 
is adopted by the \textit{pattern-based} methods, where the attention matrix is sparsified with predefined or learnable patterns. \cite{qiu2019blockwise} chunks the input sequence into fixed blocks, so that the complexity is reduced from $N^2$ to $B^2$ ($B$ is the block size smaller than $N$). Alternatively, \cite{liu2018generating} downsamples the input into fixed length. Instead of adjusting the sequence length, \cite{child2019generating,longformer} make use of strided/dilated patterns to compute attention at fixed intervals. Compared with these methods with one fixed pattern, the combination of multiple patterns can diversify the coverage of attention access \cite{tay2020efficient}. For example, \cite{ho2019axial} calculates the attention along every axis of the input, and \cite{child2019generating} aggregates the attention from both strided and local patterns. To further improve the quality of patterns, learning patterns in a data-driven manner is becoming a new trend \cite{Reformer,roy2021efficient,tay2020sparse}. The benefit over fixed patterns is that learnable ones are able to more precisely cluster input tokens based on token relevance while still maintaining the efficiency \cite{tay2020efficient}. In summary, pattern-based methods can enhance the efficiency by sparsifying the Softmax attention. However, they still have the quadratic complexity, and it will increase with the input length becoming larger. In addition to that, they are relatively complicated to reproduce in practice.

Another general category of efficient Transformers is based on \textit{kernels} (or \textit{kernel functions}), which aims to reduce the complexity from quadratic to linear. By this means, Softmax can be re-written with other forms to avoid the $N\times N$ attention matrix \cite{tay2020efficient}. To achieve this, \cite{wang2020linformer} assumes a low-rank prior within the $N\times N$ structure, and transforms the \textit{Key} and \textit{Value} into a lower dimension with extra projection layers. \cite{peng2021random} approximates Softmax with the production of a series of Gaussian kernels, and alternatively, \cite{ch2020rethinking} uses Haar kernels instead. The PerFormer \cite{2020masked} employs orthogonal random features to generate random kernels. The Linear Transformer \cite{katharopoulos2020transformers} makes use of the associative property of matrix products, and reformulates the similarity function with kernel decomposition. The cosFormer \cite{zhen2022cosformer} follows this kernelization strategy, and utilizes ReLU \cite{agarap2018deep} as the kernel function with an additional \textit{cosine} reweighting mechanism. Generally, kernel-based methods can linearize the complexity and effectively enhance the efficiency. Moreover, they are easier to implement than pattern-based approaches.

\subsection{Neural architecture search (NAS)}
NAS \cite{elsken2019neural,liu2021survey} aims to automatically find the most appropriate network architecture, and has been extensively applied to Computer Vision \cite{chen2021glit,gong2021nasvit,ding2021hr,chen2021autoformer} and Natural Language Processing \cite{so2019evolved,guan2021autoattend,wang2020textnas,ranknas,hat}. The core of NAS is to design a suitable search space, rank all candidate architectures generated from the space, and find the optimal structure. For Transformers in NLP, the Evolved Transformer (ET) \cite{so2019evolved} is the pioneering work that applies NAS into Transformer architecture search. It defines a large-scale search space, which covers the components in the input, normalization, layers, the output dimension, activations, \etc An evolution algorithm \cite{real2019regularized} is adopted to make architecture selection more stable and robust, \ig, consistently selecting the most promising architecture based on the fitness. Although ET can find a better and more efficient structure, its computational costs are still very large. This is because the algorithm has to cover all the search features. Besides, training the evolution process is also time-consuming.

Subsequent works focus on improving the search algorithm and/or pruning the search space. \cite{zhao2021memory} employs the differentiable architecture search (DARTS) \cite{liu2018darts} that includes all the operations into a node (\ig, constituting a supernet) and leverages Softmax for the specific choice. This method relaxes the need of training each candidate network separately. \cite{tsai2020finding} decomposes the Transformer structure into smaller components, and introduces an one-shot search algorithm based on sampling. HAT \cite{hat} introduces a hardware-aware constraint for accelerating the search process. RankNAS \cite{ranknas} treats NAS as a pairwise ranking problem, which significantly speeds up the search process. 

Pruning the search space, \ig, only retaining the most important search features, is another approach to reducing the search costs. TextNAS \cite{wang2020textnas} specifies a tailored search space for text representation, which consists of convolutional, recurrent, pooling, and self-attention layers. The Primer \cite{so2021primer} modifies the search space with squaring ReLU activations and a depth-wise convolutional layers after the \textit{Query}, \textit{Key}, and \textit{Value}. AutoAttend \cite{guan2021autoattend} specially searches the connections between QKV. RankNAS \cite{ranknas} proposes a feature selection algorithm that measures the importance of each feature and sorts out the most useful ones for further search. In summary, search space pruning can effectively improve the search efficiency and maintain good performance.

\section{NAS on Efficient Transformers}
In this section, we first give a brief review of the preliminary knowledge of the cosFormer \cite{zhen2022cosformer} and RankNAS \cite{ranknas}. After that, we specify the main steps of applying NAS to the efficient Transformer search and discuss the general results. Finally, we introduce the proposed idea of using mixed attention.

\subsection{Preliminary knowledge}
\label{sec:pre}
\subsubsection{cosFormer}

Given the Query $\mathbf{Q}\in \mathbb{R}^{N\times d}$, Key $\mathbf{K}\in \mathbb{R}^{N\times d}$, and Value $\mathbf{V}\in \mathbb{R}^{N\times d}$ ($d$ is the feature dimension of each head), the standard Transformer \cite{vaswani2017attention} produces the output $\mathbf{O}\in \mathbb{R}^{N\times d}$ by multiplying the normalized $\mathbf{Q}\mathbf{K}$ attention to $\mathbf{V}$, \ig,
\begin{equation}
    \mathbf{O}=\rho (\mathbf{Q}\mathbf{K}^{\mathrm{T} })\mathbf{V},
\end{equation}
\noindent
where $\rho$ is the similarity function, \eg, Softmax in the Transformer. The calculation of $\mathbf{Q}\mathbf{K}^{\mathrm{T} }\in \mathbb{R} ^{N\times N}$ leads to the quadratic complexity with respect to the sequence length $N$ (see Fig.~\ref{fig:attn_illus}(a)).

\begin{figure}[tb]
\centering
\includegraphics[width=\linewidth]{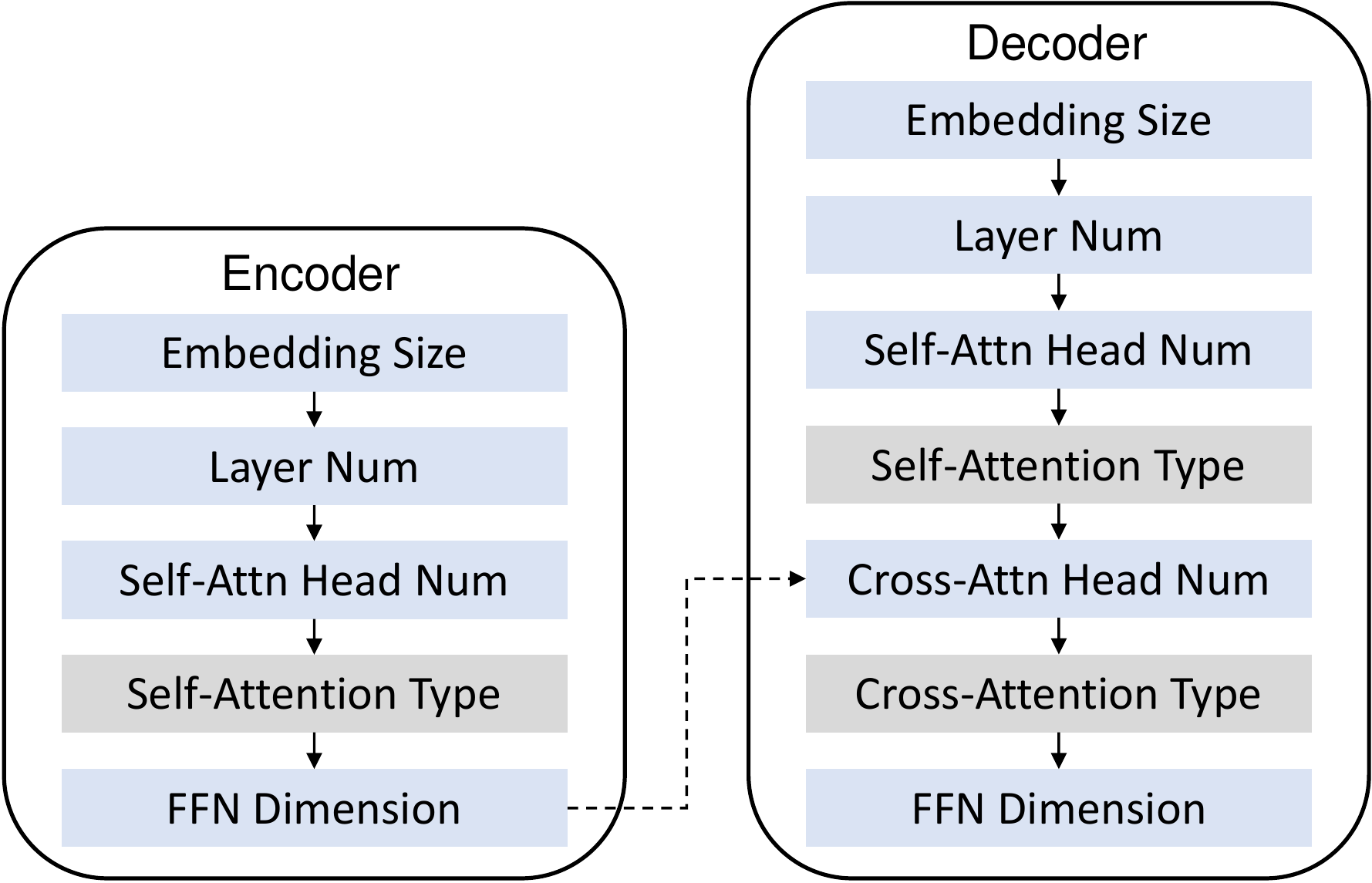}\\
%\end{tabular}
%\vspace{-6pt}
%\vspace{-12pt}
\caption{Search space in this paper. \textnormal{The blue blocks are primitive features in Transformer search, and the gray ones denote the proposed attention type search. }}
\vspace{-6pt}
\label{fig:pipeline}
\end{figure}

Linearizing attention can reduce the complexity from quadratic to linear \cite{katharopoulos2020transformers,zhen2022cosformer}. This is achieved by decomposing the similarity function $\rho$ and multiplying $\mathbf{K}$ and $\mathbf{V}$ first, \ig,
\begin{equation}
    \rho(\mathbf{Q}\mathbf{K}^\mathrm{T})\mathbf{V}=(\varphi(\mathbf{Q})\varphi (\mathbf{K})^\mathrm{T} )\mathbf{V}= \varphi(\mathbf{Q})(\varphi (\mathbf{K})^\mathrm{T} \mathbf{V}),
\label{eq:linear}
\end{equation}
\noindent
where $\varphi$ is a kernel function that transforms $\mathbf{Q}$ and $\mathbf{K}$ into hidden representations. In this case, the complexity $O(N^2d)$ lowers to $O(Nd^2)$. Generally we always have $N\gg d$, so $O(N^2d)\approx O(N^2)$ and $O(Nd^2)\approx O(N)$, yielding the linear complexity.

In essence, it is recognized that the similarity function should be \textit{non-negative} and \textit{non-linear} \cite{zhen2022cosformer,katharopoulos2020transformers}. The non-negativity strengthens the positively correlated features, and \cite{zhen2022cosformer} achieves this by directly utilizing the ReLU function \cite{agarap2018deep}, \ig, $\mathrm{ReLU}(x)=\mathrm{max}(0,x)$. The non-linearity concentrates the most relevant features to be aggregated and stabilizes training \cite{gao2017properties,aueb2016one}. To this end, \cite{zhen2022cosformer} further introduces a \textit{cosine}-based reweighting mechanism such that nearby tokens are encouraged to have higher attention connection. The entire formulation is 
\begin{equation}
    \mathbf{O}_i =\sum_{j=1}^{N} (\mathrm{ReLU}(\mathbf{Q}_i) \mathrm{ReLU}(\mathbf{K}_j)^\mathrm{T}cos(\frac{\pi }{2}\cdot \frac{i-j}{N}  ) )\mathbf{V}_j.
\label{eq:cos}
\end{equation}
\noindent
Here for simplicity, we ignore the normalization term. Eq.~\ref{eq:cos} can be re-written as the linear form in Eq.~\ref{eq:linear}, and we refer readers to the original paper \cite{zhen2022cosformer} for more details.

\subsubsection{RankNAS }
Conventional NAS methods \cite{so2019evolved,guan2021autoattend,wang2020textnas} have to evaluate the performance of every candidate network for ranking, which is time-consuming especially when the search space is large. The recently-proposed RankNAS \cite{ranknas} can effectively reduce training costs with the following three distinctions:
\begin{enumerate}
    \item \textbf{Pairwise ranking.} Instead of sorting all candidate architectures, RankNAS treats NAS as a pairwise ranking problem. That is, a binary classification is conducted to each candidate pair, and the label is simplified into \textit{correctly ordered} or \textit{incorrectly ordered} as per their estimated performance.
    \item \textbf{Search space pruning. }RankNAS prunes the search space by only containing the most important features that have large impact to the performance.
    \item \textbf{Hardware-aware search. }In addition to the standard search based on losses, RankNAS also leverages a hardware constraint, \ig, estimating the latency and discarding the fastest and slowest 10\% models.  
\end{enumerate}

\noindent
Considering the good efficiency of RankNAS, we take it as our NAS framework. More technical details can be found in \cite{ranknas}.

\subsection{RankNAS on cosFormer}

% \subsubsection{Primary tasks} We do the search on two primary tasks, \ig, machine translation (WMT'14 En-De \cite{wmt}) and image classification (CIFAR-10 \cite{cifar}).

\subsubsection{Search space}

\begin{table}[tb]
\small
\setlength{\tabcolsep}{0.1mm}{
\begin{center}
%\vspace{-5pt}
\begin{tabular}{l|l|c} 
\hline
\hline
\multirow{10}{2.7cm}{\centering \textbf{\footnotesize Primitive features} \\ \footnotesize (from RankNAS \cite{ranknas})} &Enc Layer Num   &6\\ \cline{2-3}
&Enc Emb Dim  &768, 1024  \\ \cline{2-3}
&Enc FFN Dim  &2048, 3072, 4096, 5120   \\\cline{2-3}
&Enc Head Num &4, 8, 16\\\cline{2-3}
&Dec Layer Num &1, 2, 3, 4, 5, 6\\\cline{2-3}
&Dec Emb Dim &768, 1024\\\cline{2-3}
&Dec FFN Dim &2048, 3072, 4096, 5120\\\cline{2-3}
&Dec Head Num &4, 8, 16\\\cline{2-3}
&En-De Head Num &4, 8, 16\\\cline{2-3}
&En-De Connect &1, 2, 3\\\cline{2-3}
\hline
\multirow{3}{2.7cm}{\centering \textbf{\footnotesize Proposed feature}} 
&Enc Self-Attn  & Softmax, Linear  \\\cline{2-3}
&Dec Self-Attn  & Softmax, Linear  \\\cline{2-3}
&En-De Cross-Attn & Softmax, Linear\\\cline{2-3}
\hline\hline
\end{tabular}
\end{center}}
\vspace{-15pt}
\caption{Search space for WMT'14 En-De translation. }
\vspace{-10pt}
\label{tab:mt_space}
\end{table}

Our primitive search space is adopted from RankNAS, which contains fundamental features including the embedding size, the number of encoder/decoder layers, the number of heads, and the dimension of the feed-forward network (see Fig.~\ref{fig:pipeline}). We also make corresponding modification to the space as per each task. Detailed definition is introduced in Section \ref{sec:setting}.

\subsubsection{Main steps }We replace all Softmax with the cosFormer attention, and follow RankNAS \cite{ranknas} for the search process. A supernet containing all possible architectures is firstly trained. Afterwards, the loss and latency data are separately collected and ranked. Based on the loss and latency constraints, the most important features are selected, \ig, search space pruning. The evolution algorithm is performed on the refined search space, and the optimal architecture is sorted out. Finally, we re-train the optimal network from scratch. 

\subsubsection{Discussion }

For comparison, we also repeat the above steps to the standard Transformer \cite{vaswani2017attention} only with the Softmax attention. For simplicity, we denote the optimal architecture of the cosFormer as ``cosFormer$*$'', and that of the Transformer as ``Transformer$*$''. From Fig.~\ref{fig:intro}, we find that on both tasks, the cosFormer$*$ presents better efficiency than the Transformer$*$ (smaller FlOPS) with a comparable parameter scale, but the general accuracy is less competitive. We name this phenomenon as the \textit{imbalance between accuracy and efficiency}, which has also been observed in other efficient Transformers \cite{tay2020sparse,katharopoulos2020transformers,Reformer,wang2020linformer,2020masked}.

In terms of the attention itself, the advantage of Softmax in accuracy is associated with its ability in \textit{``imposing a categorical distribution constraint on the query-context relevance scores''} \cite{zhang2021sparse}. This constraint has two essential properties: (1) \textbf{dense coverage}, \ig, attending every feature to the query \cite{vaswani2017attention}; and (2) \textbf{non-linear concentration}, \ig, larger weights are only assigned to more relevant features \cite{gao2017properties,aueb2016one,zhen2022cosformer}. In spite of the useful properties, we should also notice that the primary limitation of Softmax is its high computational complexity, as analyzed in Section \ref{sec:pre}.

\begin{table}[t]
\small
\setlength{\tabcolsep}{0.06mm}{
\begin{center}
%\vspace{-5pt}
\begin{tabular}{l|l|c} 
\hline
\hline
\multirow{4}{2.7cm}{\centering \textbf{\footnotesize Primitive features} \\ \footnotesize (from \cite{chen2021autoformer})} &Enc Layer Num   &12, 13, 14\\ \cline{2-3}
&Enc Emb Dim  &320. 384, 448 \\ \cline{2-3}
&Enc FFN Dim  &672, 896, 1344, 1568, 1792   \\\cline{2-3}
&Enc Head Num &5, 6, 7\\\cline{2-3}
\hline
\multirow{1}{2.7cm}{\centering \textbf{\footnotesize Proposed feature}} 
&Enc Self-Attn  & Softmax, Linear  \\\cline{2-3}
\hline\hline
\end{tabular}
\end{center}}
\vspace{-15pt}
\caption{Search space for CIFAR-10 classification.}
\vspace{-10pt}
\label{tab:ic_space}
\end{table}

To reduce the computational costs, efficient Transformers either sparsify the Softmax matrix (pattern-based) or replace it with other kernel functions (kernel-based). Although pattern-based methods retain the Softmax formulation, they normally utilize fixed patterns, which are less comprehensive, flexible and robust \cite{zhang2021sparse}. Kernel approaches take all features into account, but the non-linearity is less powerful in concentrating relevant features, \eg, the cosFormer \cite{zhen2022cosformer} focuses more on locality so that the relevant information in long-range distance may not be well aggregated. These efficient approximations cannot fully cover the aforementioned two properties of Softmax, so their accuracy may be negatively affected. Only a few models \cite{zhen2022cosformer,zaheer2020big,zhang2021sparse} have occasionally achieved improved accuracy over the Transformer \cite{vaswani2017attention} on some specific tasks or settings, but their performance is not consistently better in all cases. Hence, it is still challenging for efficient Transformers to approximate Softmax effectively and reduce the performance imbalance.

\subsection{Mixed attention as a new search feature}
Considering that Softmax attention and linear attention have their own distinctions but cannot simultaneously balance accuracy and efficiency well, we propose to employ the two types of attention in a mixed way. That is, Softmax attention or linear attention only occurs in certain layers, and we let the model automatically determine their layer position with NAS. 
To this end, we define a new search space, \ig, the attention type, which consists of Softmax attention and the linear attention (\ig, cosFormer attention \cite{zhen2022cosformer}). We repeat the main steps after incorporating the new search space, and name the optimal architecture as ``mFormer''. Fig.~\ref{fig:intro} demonstrates that mFormer has better balance between accuracy and efficiency on both image classification and machine translation tasks. Detailed evaluation can be found in the next section.

It should be noted that in Transformer NAS, defining a new search feature normally requires imposing extra constraints to prevent it from introducing redundant or irrelevant search content \cite{guan2021autoattend,so2019evolved}. By contrast, our attention option (only containing two types) can be directly embedded into the search space without the need of considering any constraint. This is beneficial for the convenient implementation and deployment.

\section{Experiments}
% In this section, we first compare the optimal architectures of the cosFormer and standard Transformer. With the comparison, we provide insights on how to design a customized, effective, and efficient network structure for efficient Transformers. After that, we stress the advantage of the proposed mixed attention in balancing accuracy and efficiency through extensive experiments.

\subsection{Experimental settings}
\label{sec:setting}
\subsubsection{Datasets }\textbf{Machine translation: }We use the WMT'14 En-De dataset \cite{wmt} containing 4.5M pairs of sentences. The training, validation, and test split is identical to that of \cite{hat,ranknas}, \ig, WMT'16 for training, Newstest2013 for validation, and Newstest2014 for test. \textbf{Image classification: }We leverage the CIFAR-10 dataset \cite{cifar}, which consists of 60K $32\times32$ images in 10 classes. We use the official split for training (50K) and test (10K).

\subsubsection{Search space definition }\textbf{Machine translation: }The search space for WMT'14 En-De is adopted from RankNAS \cite{ranknas}, as listed in Table \ref{tab:mt_space}. The number of encoder layers is set as 6, which is consistent with \cite{ranknas,hat}. ``En-De Connect'' represents the number of encoder layers attended to the decoder, \eg, 2 means the last two encoder layers are involved in calculating the encoder-decoder cross-attention. The proposed attention search is composed of the Softmax attention and linear attention (\ig, the cosFormer attention \cite{zhen2022cosformer}). Note that in practice, we do not include the attention type search in decoder self-attention and encoder-decoder cross-attention. This is because we empirically find that employing the linear attention into decoder layers would lead to inappropriate convergence and poor performance\footnote{We refer the reader to the supplementary material for more details.}. \textbf{Image classification: }For CIFAR-10, we follow the setting in \cite{chen2021autoformer}, \ig, only using the encoder part and constructing the primitive search space in Table \ref{tab:ic_space}.

\subsubsection{Training configuration }We perform all experiments on A100 GPU cards with the same training steps as RankNAS \cite{ranknas}. On the image classification task, the CIFAR-10 images are first downsampled with convolutional layers to $14\times14$, and then flattened as an 1D vector with 256 elements \cite{chen2021autoformer}. The vector is the input to the Transformer.

\subsubsection{Evaluation metrics }For efficiency in both two tasks, we calculate the FLOPS (floating point operations per second). For accuracy, we measure BLEU (bilingual evaluation understudy) for translation with beam 4 and length penalty 0.6. The model is tested with the averaged last ten checkpoints. The accuracy of CIFAR-10 is evaluated by computing the percentage of correctly classified images. Both metrics are implemented with the source code in \cite{ranknas} and measured on a Nvidia Tesla A100 card.

\subsection{Optimal architecture comparison on machine translation}
From Table \ref{tab:ablation}, the searched architectures of the cosFormer and Transformer are more efficient while retaining comparable accuracy to the vanilla models. It validates the effectiveness of NAS. We compare optimal architectures below in detail.
\subsubsection{Network structure} The searched optimal architecture of the cosFormer \cite{zhen2022cosformer} is illustrated in Table \ref{tab:mt_search}. For comparison, we also report the searched results of the Transformer \cite{vaswani2017attention}. In spite of the identical embedding size in both encoders and decoders, the cosFormer$*$ has a lighter structure, \eg, fewer decoder layers, smaller encoder FFN dimensions and decoder head numbers. It implies that the optimal structure of the cosFormer tends to be \textbf{``shallow and thin''}. 

%\weixuan{a little comfusing? smaller models certainly has worse performance.}

\begin{figure*}[t]
\centering
\includegraphics[width=\linewidth]{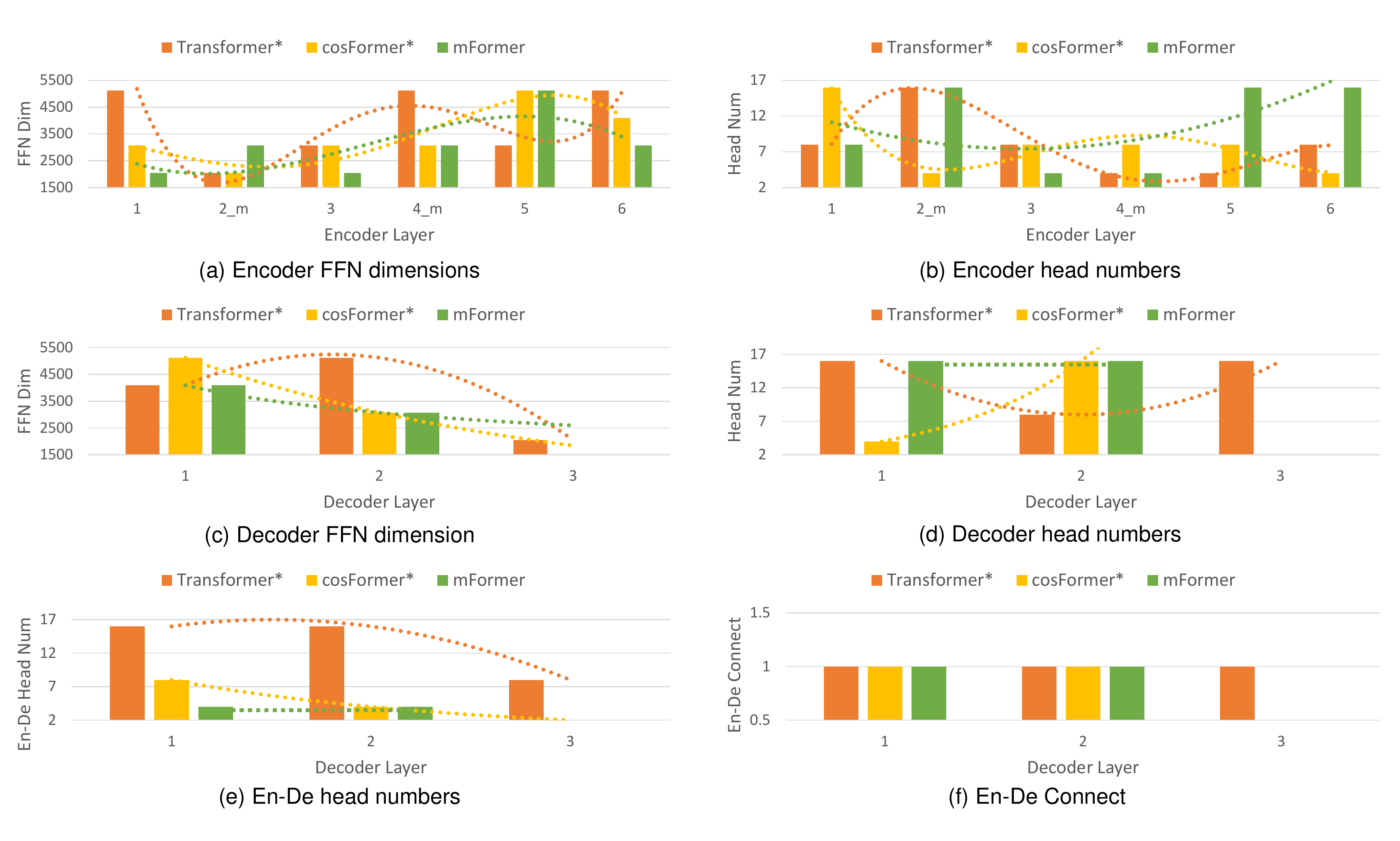}\\
%\end{tabular}
\vspace{-12pt}
%\vspace{-12pt}
\caption{Detailed searched features on WMT'14 En-De. \textnormal{Layers with linear attention are suffixed with ``\_m''. Dashed lines reflect the tendency of changes.}}
\vspace{-6pt}
\label{fig:mt_detail}
\end{figure*}

\textbf{FFN dimension.} Fig.~\ref{fig:mt_detail}(a) and (c) display the FFN dimensions in the encoder and decoder respectively. We draw the tendency curves with the polynomial fitting function for better visualization of changes. The overall FFN dimension of the cosFormer$*$ is smaller than the Transformer$*$. More specifically, larger dimensions are required in top layers of the encoder in the cosFormer$*$. Both the cosFormer$*$ and Transformer$*$ tend to have a smaller dimension in last layers of the decoder.

\begin{table*}[t]
\small
\setlength{\tabcolsep}{5.8mm}{
\begin{center}
%\vspace{-5pt}
\begin{tabular}{c|l|c|c|c} 
\hline
\hline
 && Transformer$*$ &cosFormer$*$ &mFormer (ours) \\\cline{1-5}
\multirow{10}{2.7cm}{\centering \textbf{\footnotesize Searched architectures} }&Enc Layer Num   &3&6 &6\\ \cline{2-5}
&Enc Emb Dim  &1024  &1024 &1024\\ \cline{2-5}
&Enc FFN Dim AVG   &3925  &3413 &3072 \\\cline{2-5}
&Enc Head Num AVG &8 &8 &10.67\\\cline{2-5}
&Dec Layer Num &3&2&2\\\cline{2-5}
&Dec Emb Dim &1024&1024&1024\\\cline{2-5}
&Dec FFN Dim AVG &3755&4096&3584\\\cline{2-5}
&Dec Head Num AVG &13&10&16\\\cline{2-5}
&En-De Head Num AVG &13&6&4\\\cline{2-5}
&En-De Connect AVG &1&1&1\\\cline{2-5}

\hline\hline
\multirow{3}{2.7cm}{\centering \textbf{\footnotesize Searched results} } &\#Parameters (M) & 92.38&79.98 &75.59 \\ \cline{2-5}
&FLOPS (G) &9.86 &8.32 &8.08 \\ \cline{2-5}
&BLEU &28.34 &3.05 &25.79 \\ \hline\hline
\end{tabular}
\end{center}}
\vspace{-15pt}
\caption{Searched architectures and results on machine translation. \textnormal{``AVG'' means the average value.}}
\vspace{-10pt}
\label{tab:mt_search}
\end{table*}

\textbf{Head number.} We plot head numbers in Fig.~\ref{fig:mt_detail}(b)(d)(e). In general, the cosFormer$*$ requires a smaller number of heads. 

\textbf{En-De Connect. }From Fig.~\ref{fig:mt_detail}(f), we observe that decoders in both models only attend the last layer of the encoder, \ig, obtaining the cross information only from the last encoder block.

\textbf{Parameters.} The cosFormer$*$ has 12.6M fewer parameters than the Transformer$*$.

\subsubsection{Accuracy} On the WMT'14 En-De test set, the cosFormer$*$ presents extremely poor BLEU\footnote{It has the inappropriate convergence, and we are communicating with the authors of \cite{zhen2022cosformer} for dealing with this issue.}. It reflects the limitation of the linear attention in achieving comparable accuracy to the Transformer$*$.

We also investigate the importance of each feature to the accuracy, which is measured by RankNAS \cite{ranknas}. The top-5 ranked features are:
\begin{itemize}
     \item \textbf{cosFormer$*$: }En-De Connect, Dec Layer Num, En-De Head Num, Dec FFN Dim, Dec Head Num.
    \item \textbf{Transformer$*$: }Dec Layer Num, En-De Head Num, Dec FFN Dim, Dec Head Num, Enc FFN Dim.
   
\end{itemize}
\noindent

Hence, when designing the architecture for the efficient Transformer, we need to focus more on the encoder-decoder interaction, \eg, the number of encoder layers to attend and the cross-attention heads. Both the cosFormer$*$ and Transformer$*$ are also sensitive to the decoder layer number, but the best performance does not necessarily need the largest number of layers.

\begin{table*}[t]
\small
\setlength{\tabcolsep}{5.8mm}{
\begin{center}
%\vspace{-5pt}
\begin{tabular}{c|l|c|c|c} 
\hline
\hline
 && Transformer$*$ &cosFormer$*$ &mFormer (ours) \\\cline{1-5}
\multirow{4}{2.7cm}{\centering \textbf{\footnotesize Searched architectures} }&Enc Layer Num   &14&12 &12\\ \cline{2-5}
&Enc Emb Dim  &384  &448 &384\\ \cline{2-5}
&Enc FFN Dim AVG &1328  &1419 &1232 \\\cline{2-5}
&Enc Head Num AVG &6.86 &6.25 &6.33\\\cline{2-5}

\hline\hline
\multirow{3}{2.7cm}{\centering \textbf{\footnotesize Searched results} } &\#Parameters (M) &24.26 &24.31 &19.31 \\ \cline{2-5}
&FLOPS (G) &10.90 &9.55 &8.39 \\ \cline{2-5}

& Accuracy (\%) &95.10 &88.40 &93.59 \\ \hline\hline
\end{tabular}
\end{center}}
\vspace{-15pt}
\caption{Optimal architecture and performance comparison on CIFAR-10 image classification. \textnormal{``AVG'' means the average value.}}
\vspace{-10pt}
\label{tab:ic_search}
\end{table*}

\begin{figure*}[tb]
\centering
\includegraphics[width=\linewidth]{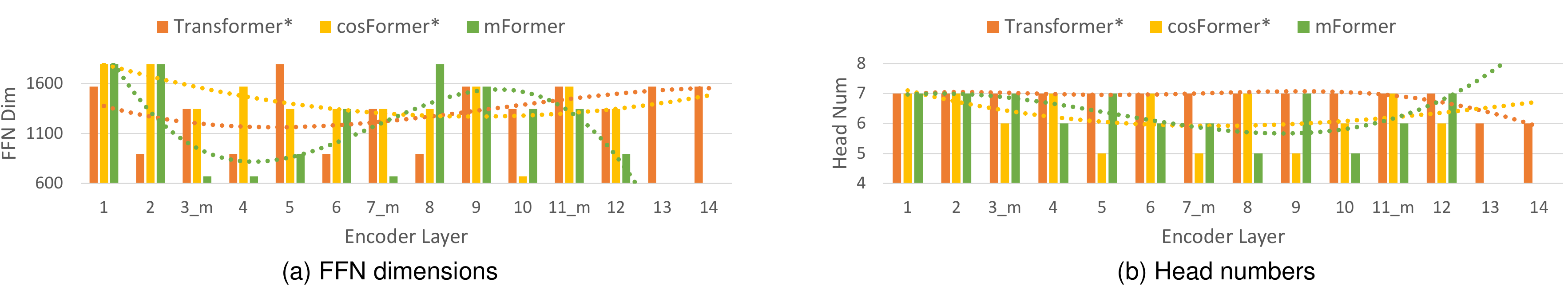}\\
%\end{tabular}
\vspace{-6pt}
%\vspace{-12pt}
\caption{Detailed searched features on image classification. \textnormal{Layers with linear attention are suffixed with ``\_m''. Dashed lines reflect the tendency of changes.}}
\vspace{-5pt}
\label{fig:ic_detail}
\end{figure*}

\subsubsection{Efficiency} Clearly, the cosFormer$*$ is more efficient, \ig, its FLOPS is 8.32G while that of the Transformer$*$ is 9.86G. This difference is mainly due to the more complex optimal structure of the Transformer. The Softmax attention also increases the computational costs, which will be detailed in Section \ref{sec:abb}. The top-5 ranked features regarding the efficiency are:
%\vspace{-12pt}
\begin{itemize}
     \item \textbf{cosFormer$*$: }Dec Layer Num, En-De Head Num, En-De Connect, Enc FFN Dim, Dec Head Num.
    \item \textbf{Transformer$*$: }Dec Layer Num, Dec Head Num, Enc Head Num, Enc FFN Dim, Dec FFN Dim.
\end{itemize}
\noindent
It is obvious that the number of decoder layers has the largest impact to the efficiency. For the efficient Transformer, the encoder-decoder interaction also plays an essential role in determining the computational costs. By contrast, the Transformer$*$'s efficiency is less sensitive to the cross interaction.

\begin{table*}[t]
\small
\setlength{\tabcolsep}{4.8mm}{
\begin{center}
%\vspace{-5pt}
\begin{tabular}{c|l|c|c|c} 
\hline
\hline
 & & \#Parameters (M) &FLOPS (G) & BLEU/Accuracy (\%)\\\cline{1-5}

 \multirow{9}{2.7cm}{\centering \textbf{\footnotesize Machine translation} }
 
&mFormer  &75.59  &8.085 &25.79\\ \cline{2-5}
&mFormer (all linear) &75.59  &8.081 &3.53 \\ \cline{2-5}
&mFormer (all Softmax) &75.59 &8.09 &27.95\\ \cline{2-5}
&Vanilla cosFormer \cite{zhen2022cosformer}  &213.00  &13.39 &Null \\\cline{2-5}
&cosFormer$*$  &79.78  &8.32 &3.05 \\\cline{2-5}
&cosFormer$*$ (all Softmax) &79.78 &8.34 &27.83\\\cline{2-5}
&Vanilla Transformer \cite{vaswani2017attention}  &213.00&12.7 &28.4\\ \cline{2-5}
&Transformer$*$   &92.38&9.86 &28.34\\ \cline{2-5}
&Transformer$*$ (all linear)  &92.38  &9.84 &3.47\\ \cline{2-5}

\hline\hline

\multirow{9}{2.7cm}{\centering \textbf{\footnotesize Image classification} }
&mFormer  &19.31  &8.39 &93.59\\ \cline{2-5}
&mFormer (all linear) &19.23 &7.55 &83.49 \\\cline{2-5}
&mFormer (all Softmax) &19.33 &8.67 &94.35\\\cline{2-5}
&Vanilla cosFormer \cite{zhen2022cosformer}  &19.34  &7.60 &87.13 \\\cline{2-5}
&cosFormer$*$  &24.31  &9.55 &88.40 \\\cline{2-5}
&cosFormer$*$ (all Softmax) &24.42 &10.66 &94.80\\\cline{2-5}
&Vanilla Transformer \cite{vaswani2017attention}  &19.31&8.07 &88.70\\ \cline{2-5}
&Transformer$*$   &24.26&10.90 &95.10\\ \cline{2-5}
&Transformer$*$ (all linear)  &24.14  &9.48 &81.82\\ \cline{2-5}

\hline\hline

\end{tabular}
\end{center}}
\vspace{-15pt}
\caption{Ablation study on attention types.}
\vspace{-10pt}
\label{tab:ablation}
\end{table*}

%\vspace{-8pt}

\subsubsection{Summary }We summarize empirical hints to design an appropriate architecture for the efficient Transformer on machine translation: (1) Fewer decoder layers; (2) Focusing more on the encoder-decoder interaction. The encoder layers to attend and the number of encoder-decoder heads are not necessarily too large; (3) The encoder FFN dimension is more important to the efficiency while the decoder FFN dimension is more important to the accuracy.

\subsection{Optimal architecture comparison on image classification}
The optimal structures of the cosFormer and Transformer significantly outperform their vanilla versions in accuracy (see Table \ref{tab:ablation}, which further verifies the necessity and usefulness of NAS. The FLOPS is larger because vanilla models only have 6 encoder layers. We compare optimal structures in the following.

\subsubsection{Network structure} Table \ref{tab:ic_search} shows the optimal architectures of the cosFormer \cite{zhen2022cosformer} and Transformer \cite{vaswani2017attention}. Clearly, the Transformer$*$ has more layers, a smaller embedding size, and a smaller averaged FFN dimension than the cosFormer$*$. From a general view, the cosFormer prefers a \textbf{``shallow and wide''} structure while the optimal architecture of the Transformer tends to be \textbf{``deep and thin''}. 

\textbf{FFN dimension.} Fig.~\ref{fig:ic_detail}(a) displays the FFN dimensions in different layers. From the tendency curve, we observe that the cosFormer$*$ has a relatively smaller FFN dimension in intermediate layers. Differently, the dimension is smaller in early layers in the Transformer$*$.

\textbf{Head number.} The head numbers in encoder layers are plotted in Fig.~\ref{fig:ic_detail}(b). The cosFormer$*$ generally has more heads in early and top layers. By contrast, the head numbers in the Transformer$*$ do not vary too much and their values are large. 

\textbf{Parameters. }The cosFormer$*$ and the Transformer$*$ have a similar parameter scale, \ig, $\sim$24M. 

\subsubsection{Accuracy} On the CIFAR-10 test set, the cosFormer$*$ achieves the 88.4\% accuracy, which is surpassed by the Transformer$*$ (95.10\%) by around 7.6\%.

According to the importance to the accuracy, the features ranked by RankNAS \cite{ranknas} are:
\begin{compactitem}
     \item \textbf{cosFormer$*$: }Layer Num, FFN Dim, Emb Dim, Head Num.
    \item \textbf{Transformer$*$: }FFN Dim, Head Num, Emb Dim, Layer Num.
   
\end{compactitem}
\noindent
Clearly, the number of layers has the largest impact to the accuracy of the cosFormer. Similar to the case in machine translation, the cosFormer$*$ does not select the greatest layer number.

\subsubsection{Efficiency} The FLOPS of the cosFormer$*$ is 7.01G, which is significantly better than the Transformer$*$ (10.9G) by 36\%. It reveals the advantage of the cosFormer's linear attention in efficiency. The important features regarding the efficiency are:
\begin{compactitem}
     \item \textbf{cosFormer$*$: }Layer Num, Emb Dim, Head Num, FFN Dim.
    \item \textbf{Transformer$*$: }Layer Num, FFN Dim, Head Num, Emb Dim.
   
\end{compactitem}
\noindent
Similar to the conclusion in machine translation, the layer number in image classification is also the most related to the efficiency. The efficient model has the preference of using fewer layers, which further reduces the computational burden.

\subsubsection{Summary} We give a brief summary of how to appropriately design the efficient Transformer in image classification: (1) Fewer layers; (2) A smaller FFN dimension in intermediate layers with a slightly larger embedding size; (3) A smaller head number in intermediate layers.

\subsection{Our results using mixed attention}
\subsubsection{Machine translation} Table \ref{tab:mt_search} shows the optimal structure and the performance of our mFormer. In general, the mFormer is \textbf{``shallow and thin''}, which has the fewest parameters and smallest FLOPS among the three optimal models. The linear attention exists in Layer 2 and 4 in the encoder, and all the other encoder/decoder layers retain the Softmax attention. In mFormer, the FFN dimensions in the encoder undergo an ``up-and-down'' change, \ig, early layers rely on smaller FFN sizes. The number of heads in the encoder tends to be smaller in intermediate layers. The primary features in the decoder of the mFormer consistently have the no larger quantity than the other two, except that the head number in the first decoder layer is larger than that of the cosFormer$*$.

In terms of the performance, the mFormer achieves comparable BLEU to the Transformer$*$ with 18\% fewer parameters and 18\% smaller FLOPS. It also significantly outperforms the cosFormer$*$ in accuracy with slightly better efficiency. The results indicate that the mixed use of attention can effectively facilitate the reduction of the imbalance between accuracy and efficiency.

\subsubsection{Image classification} We display the optimal architecture of the proposed mFormer in Table \ref{tab:ic_search}. Intuitively, our structure is \textbf{"shallow and thin"}, which is consistent with the observation in machine translation. This is also reflected by the parameter scale, \ig, we have around 20\% fewer parameters than the other two. The linear attention exists in Layer 3, 7 and 11 (accounting for 25\%), and the remaining attention types in other layers are all Softmax. The FFN dimensions are large in early layers, and has a ``down-up-down'' tendency in subsequent layers. The changes of head numbers seem to be inversely proportional to the FFN dimension, \ig, larger FFN dimensions normally correspond to smaller head numbers.  

Remarkably, the mFormer outperforms the cosFormer$*$ to a large margin in accuracy (by 5.19 in percentage) while maintaining good efficiency, \ig, it only has 1.38G more FLOPS than the cosFormer$*$. Moreover, it achieves comparable performance to the Transformer$*$ in accuracy (the latter surpasses ours only by 1.51 in percentage) but is significantly more efficient (our FLOPS is smaller than the Transformer$*$ by 2.51G, which is $\sim$23\%). It demonstrates that the use of mixed attention can achieve a better balance between accuracy and efficiency on the image classification task.

% More layers will inevitably increase the general model complexity. For example, they have to consume larger memory because all layers need to store their corresponding activations for back-propagation \cite{Reformer}. Also, the extra use of Softmax leads to larger computational costs due to its quadratic nature. With the hardware constraint from RankNAS \cite{ranknas}, the Transformer$*$ leans to having a relatively smaller embedding size and FFN dimension to prevent the computational complexity from growing too high. The circumstance in the cosFormer$*$ is slightly opposite. Although it has fewer layers, its embedding and FFN sizes are larger to provide sufficient parameters and maintain stable performance (the loss constraint imposed by RankNAS \cite{ranknas}).  

\subsection{Ablation study on attention types}
\label{sec:abb}
Since the attention search is newly introduced in the NAS framework, we specially study the influence of each attention type in optimal architectures. This is to validate (1) the effectiveness of the proposed mixed attention in balancing accuracy and efficiency, and (2) the inherent difference between Softmax and linear attention in performance. The results on machine translation and image classification are reported in Table \ref{tab:ablation}.

\subsubsection{All Softmax attention }
We first uniform the mixed attention and linear attention in the optimal architectures of the mFormer and cosFormer$*$ with Softmax. For our mFormer, the accuracy is slightly improved but the FLOPS becomes larger, indicating that the better accuracy is always along with the sacrifice in efficiency. This performance imbalance is more obvious when we replace all the linear attention in the cosFormer$*$ with Softmax, where the accuracy is significantly enhanced with much degraded computational efficiency. Notably, our model is able to achieve comparable accuracy to the performance with all Softmax and keep satisfactory efficiency without dropping too much.  

\subsubsection{All linear attention }In this step, we replace the mixed attention in the mFormer and Softmax attention in the Transformer$*$ uniformly with the linear attention. We notice a significant performance drop in accuracy on both tasks. It demonstrates the less competitive ability of the linear attention in yielding accurate results. However, after the replacement, the efficiency gain is remarkable, which further indicates the advantage of the efficient Transformer in reducing computational costs. The mixed use of attention is beneficial for achieving both comparable accuracy and efficiency.

%\vspace{-2pt}

\section{Conclusion}
In this paper, we utilize NAS to find the optimal architecture of efficient Transformers (\ig, the cosFormer \cite{zhen2022cosformer}). The searched architectures reveal that the optimal structures of the efficient Transformer are relatively lighter than those of the standard Transformer, \eg, the reduced parameter scale and improved FLOPS. This provides useful insights to the community for the appropriate design of efficient Transformers. However, the general accuracy of efficient models is less competitive. Based on this observation, we propose a novel usage of attention, \ig, employing the linear attention and Softmax in a mixed manner in each layer. The searched optimal architecture presents comparable accuracy to the Transformer and maintains as good efficiency as the efficient Transformer. The proposed method supplies a new direction in the Transformer study, \ig, taking advantage of both Softmax and linear attention. In our future work, we will study the mixed attention on large-scale pretrain models as well as other downstream tasks.
% \begin{acks}
% \end{acks}

% \bibliographystyle{ACM-Reference-Format}

{\small
\bibliographystyle{ieee_fullname}
\bibliography{main}
}

\clearpage
\title{\textbf{Supplementary Material}}
\maketitle

\section{Training Details}
We specify training details of machine translation and image classification tasks below. For those not listed, we use the default setting in RankNAS \cite{ranknas}.
\subsection{Machine translation}
\begin{compactitem}
    \item max relative length: 8
    \item optimizer: adam
    \item adam betas: (0.9, 0.998)
    \item weight decay: 0.1
    \item max tokens: 4400
    \item criterion: label smoothed cross entropy
    \item label smoothing: 0.1
    \item min learning rate: $10^{-9}$
    \item max update: 200000
    %\item max sentences: 64
    \item warmup updates: 4000
    \item lr scheduler: inverse\_sqrt
    \item lr: 0.0007
    \item warmup init lr: $10^{-7}$
    \item max lr: $1$
\end{compactitem}

\subsection{Image classification}

\begin{compactitem}
    \item max relative length: 14
    \item optimizer: adam
    \item adam betas: (0.9, 0.998)
    \item weight decay: 0.05
    \item max tokens: 100000
    \item criterion: label smoothed cross entropy
    \item label smoothing: 0.1
    \item min learning rate: $10^{-7}$
    \item max update: 140000
    \item max sentences: 64
    \item warmup updates: 976
    \item lr scheduler: cosine
    \item lr: $10^{-6}$
    \item warmup init lr: $10^{-6}$
    \item max lr: $10^{-3}$
\end{compactitem}

\section{Feature Importance in mFormer}
With RankNAS \cite{ranknas}, we can know the importance of each feature when searching the proposed mFormer. 

\subsection{Machine translation}
The top 5 features regarding the accuracy are: Dec Layer Num, En-De Head Num, En-De Head Num, Enc Attn Type, and Dec FFN Dim. Similar to the Transformer$*$, the most important feature is the number of decoder layers. In our optimal structure, this number is 2, which indicates that the best performance does not necessarily need many layers. Besides, the encoder-decoder interaction also has large impact to the accuracy, which requires careful design. The proposed attention type is important as well, further validating the effectiveness of attention search. 

The top 5 features regarding the efficiency are: Dec Layer Num, Dec Head Num, En-De Connect, En-De Head Num, and Enc FFN Dim. Again, the number of decoder layers has the largest impact to the efficiency. Note that the proposed attention type is not contained here, which is because it only exists in encoders with 6 layers and the efficiency gain is thus not very large.

\subsection{Image classification}
The top 5 features regarding the accuracy are: Enc Attn Type, Enc Layer Num, Enc FFN Dim, Enc Head Num, and Enc Embed Dim. The top 5 features regarding the efficiency are: Enc Layer Num, Enc Attn Type, Enc Embed Dim, Enc Head Num, and Enc FFN Dim.

Clearly, on this task, our attention type plays a crucial role in both accuracy and efficiency, which well verifies the effectiveness of the mixed use of attention.

\section{Linear Attention in Decoders}
On the machine translation task, when we put the linear attention in decoder layers, the BLEU values are all 0 (corresponding to three cases: 1) only decoder self attention, 2) only encoder-decoder attention, and 3) both). We also find that the output sentences have the ``repeater'' problem, \ig, repeating a single word for many times. We have not found the specific reason for this issue, and we are communicating it with the authors in \cite{zhen2022cosformer}. We infer that the potential reason is the locality in the cosFormer attention \cite{zhen2022cosformer} may be more suitable for feature fusion in encoders.

%%%%%%%%% REFERENCES

\end{document}